\documentclass[twocolumn]{el-author}

\usepackage{subfigure}
\usepackage{times}
\usepackage{epsfig}
\usepackage{graphicx}
\usepackage{amsmath}
\usepackage{amssymb} 
\usepackage{threeparttable}
\usepackage{booktabs}
\usepackage{multirow}

\newcommand{\ua}{\uparrow}
\newcommand{\nc}{\newcommand}
\nc{\da}{\downarrow} \nc{\hc}{\hat{c}} \nc{\hS}{\hat{S}}
\nc{\bra}{\langle} \nc{\ket}{\rangle} \nc{\eq}{equation (\ref}
\nc{\h}{\hat} \nc{\hT}{\h{T}}\nc{\be}{\begin{eqnarray}}
\nc{\ee}{\end{eqnarray}}\nc{\rd}{\textrm{d}}\nc{\e}{eqnarray}\nc{\hR}{\hat{R}}\nc{\Tr}{\mathrm{Tr}}
\nc{\tS}{\tilde{S}}\nc{\tr}{\mathrm{tr}}\nc{\8}{\infty}\nc{\lgs}{\bra\ua,\phi|}\nc{\rgs}{|\ua,\phi\ket}
\nc{\hU}{\hat{U}}\nc{\lfs}{\bra\phi|}\nc{\rfs}{|\phi\ket}\nc{\hZ}{\hat{Z}}\nc{\hd}{\hat{d}}\nc{\mD}{\mathcal{D}}
\nc{\bd}{\bar{d}}\nc{\bc}{\bar{c}}\nc{\mc}{\mathcal}\nc{\ea}{eqnarray}\nc{\mG}{\mathcal{G}}\nc{\bce}{\begin{center}}
\nc{\ece}{\end{center}}
\date{12th December 2011}

\begin{document}

\title{E2-Capsule Neural Networks for Facial Expression Recognition Using AU-Aware Attention}

\author{Shan Cao, Yuqian Yao and Gaoyun An}

\abstract{Capsule neural network is a new and popular technique in deep learning. However, the traditional capsule neural network does not extract features sufficiently before the dynamic routing between the capsules. In this paper, the one Double Enhanced Capsule Neural Network (E2-Capsnet) that uses AU-aware attention for facial expression recognition (FER) is proposed. The E2-Capsnet takes advantage of dynamic routing between the capsules, and has two enhancement modules which are beneficial for FER. The first enhancement module is the convolutional neural network with AU-aware attention, which can help focus on the active areas of the expression. The second enhancement module is the capsule neural network with multiple convolutional layers, which enhances the ability of the feature representation. Finally, squashing function is used to classify the facial expression. We demonstrate the effectiveness of E2-Capsnet on the two public benchmark datasets, RAF-DB and EmotioNet. The experimental results show that our E2-Capsnet is superior to the state-of-the-art methods. Our implementation will be publicly available online.}

\maketitle

\section{Introduction}

As an important branch of face recognition, Facial Expression Recognition (FER) is currently a hot topic in the field of pattern recognition. It has a wide range of applications, such as psychoanalysis, clinical medicine and human-computer interaction.

Most methods for FER only focus on feature representation of the whole face. There has been a little literature that considers local and global features. Fan et al. \cite{1} proposed MRE-CNN  framework for FER, which aimed to enhance the learning power of CNN models by capturing both the global and local features. Li et al. \cite{2} put forward a new DLP-CNN method that enhanced the discriminative power of deep features by preserving the locality closeness while maximizing the inter-class scatter. 

However, they neglect the relative relationship between local features. The face has a certain structure, and each part of the face has the relative relationship. Considering the capsules can capture the relationship between features of different local regions, we propose the E2-Capsnet for facial expression recognition. In our E2-Capsnet, AU-aware attention is introduced in traditional convolutional neural networks to achieve richer and more robust information, which is then encoded by dynamic routing between the capsules for better feature representation. Finally, the result of facial expression classification is achieved by squashing function. The experimental results show that our E2-Capsnet has excellent performance on RAF-DB and EmotioNet datasets.

\section{E2-Capsnet}
\begin{figure}[t]
	\centering{\includegraphics[width=80mm]{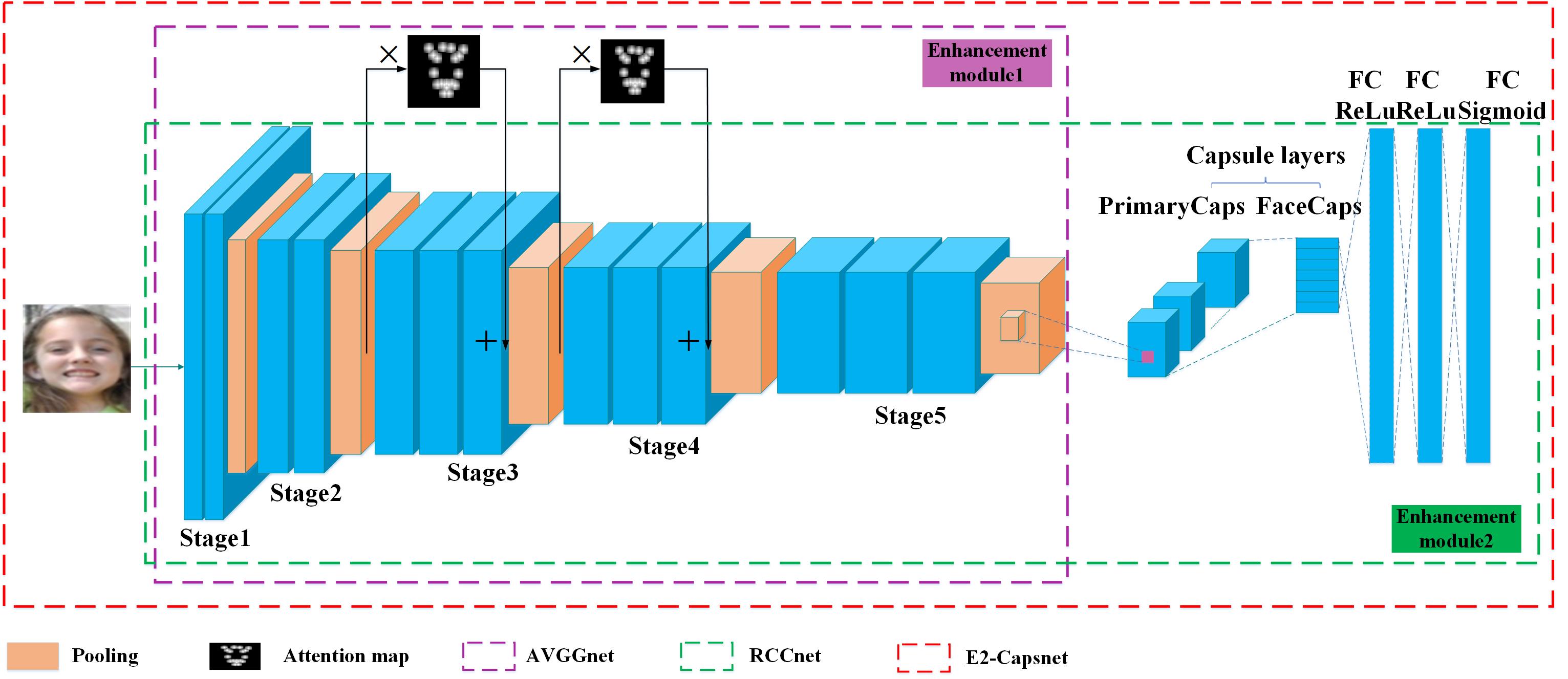}}
	
	\caption{The structure of our E2-Capsnet, consisting of two enhancement modules. E2-Capsnet takes a facial image as input and extracts rich feature maps with enhancement module1. Then the feature maps are fed to the capsule layers to be encoded. Finally, we get the result of FER by squashing function. Our E2-Capsnet is trained end-to-end.}  
	\label{fig:rpn}
\end{figure}

The structure of the entire network is shown in Fig.~\ref{fig:rpn}. Our network consists of two parts: (1) The enhanced feature maps are obtained by the enhancement module1 (as shown with the purple dashed box in Fig.~\ref{fig:rpn}) with AU-aware attention and the deep convolutional layers; (2) The enhanced feature maps are encoded through dynamic routing between capsules and then decoded by fully-connected layers (as shown with the green dashed box in Fig.~\ref{fig:rpn}). We utilize squashing function to obtain the results of facial expression recognition. 

VGG16 is chosen as the backbone network for E2-Capsnet due to its simple structure and excellent performance in object classification. For the convenience of description, the VGG16 network is described as five stages. Each stage includes multiple convolutional layers followed by a max-pooling layer. The first and second stages have two convolutional layers, and the last three stages have three convolutional layers respectively. Since we are going to get the feature maps, the last three fully connected layers are not retained.

\begin{figure}[t]

	\centering{\includegraphics[width=30mm]{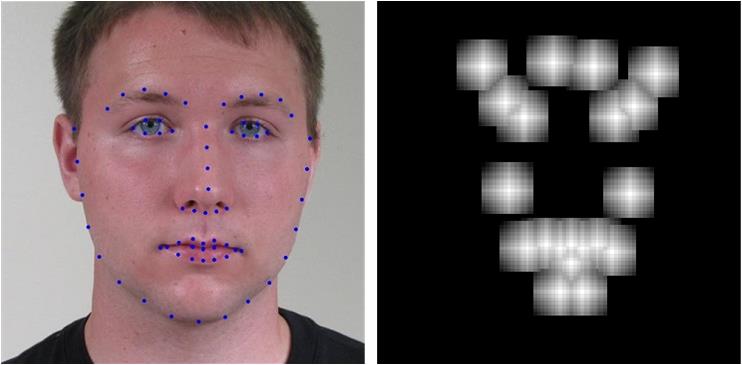}}

	\caption{Attention map generation. Left: landmarks (blue) on a face; Right: attention map of the face.}
	\label{fig:long}
	
\end{figure}
Fig.~\ref{fig:long} shows our attention map based on key facial landmarks. We refer to the generation method of attention map by Wei Li \cite{3} and make appropriate adjustments to the datasets we used. The landmarks for the key points on the face as shown with blue points in Fig.~\ref{fig:long} are needed to obtain attention maps. With the key facial landmarks, the AU centers can be obtained by shifting the scaled distance or directly using existing facial landmarks.  
The facial image is resized to $100\times100$ pixels to make sure the same scales are shared among all images. In order to make the shifting distance more adaptable to all facial images, a measurement reference is defined for the shifting distance. Inner corner distance is used as the scaled-distance, which is helpful to locate the AU centers. The nearby 7 pixels belonging to the same area for each AU center are applied in our experiments, therefore the size of each AU area is $15\times15$. Higher weight $W_a$ is assigned to the closer points to the AU center.
\begin{equation}
W_a=1-0.07d_m
\label{fig:111}
\end{equation}
where $d_m$ is the Manhattan distance to the AU center. The areas in the attention map with higher values correspond to the AU active areas in the facial images, which can be enhanced by the attention map.

As shown in Fig.~\ref{fig:rpn}, the generated attention maps are embedded in stage 3 and stage 4. The specific operation is to multiply the feature maps generated after the pooling layer of the stage 2 by the first attention map, and then parallel with the convolution of the stage 3. The convolution results of stage 3 are added to them element by element, and then are used as the input of the max-pooling layer of this stage. Likewise, the same operation is performed jointly by the attention map with the convolutional layers in stage 4. The reason why we design the attention map is that not all the areas of a facial image are equally important for FER. 

We can get $512\times7\times7$ feature maps by enhancement module1. Then the feature maps are fed to the capsule layers. Dynamic routing is performed between the PrimaryCaps layer and the FaceCaps layer. The decoder consisting of 3 fully connected layers reconstruct the face. Squashing function is used to achieve the result of FER. This nonlinear function is defined Eq.~\ref{fig:222}:
\begin{equation}
v_j=\frac{\Vert s_j \Vert ^2}{1+\Vert s_j \Vert ^2}\frac{s_j}{\Vert s_j \Vert}
\label{fig:222}
\end{equation}
where $v_j$ is the output vector of the capsule $j$, and $s_j$ is the input vector of the capsule $j$. Parameters in the network are updated by minimizing margin loss ($L_c$) and reconstruction loss ($L_r$). The expression of the loss functions are defined in Eq.~\ref{fig:777}, Eq.~\ref{fig:888} and Eq.~\ref{fig:999}:
\begin{equation}
L_c=T_cmax(0,m^+-\Vert v_c \Vert)^2+\lambda(1-T_c)max(0,\Vert v_c \Vert-m^-)^2
\label{fig:777}
\end{equation}

\begin{equation}
L_r=(x_r-x)^2
\label{fig:888}
\end{equation}

\begin{equation}
L=L_c+0.0005L_r
\label{fig:999}
\end{equation} 
where $c$ represents the classification category and $T_c$ represents the indication function of the classification. $m^+$ represents the upper boundary and $m^-$ represents the lower boundary. $x$ and $x_r$ represent the original image and the reconstructed image, respectively. 

\section{Experimental Settings}
VGG16 is adopted as the backbone network for E2-Capsnet, so all of the input images to the networks are resized to $224 \times224$. The $512\times7\times7$ feature maps are obtained by enhancement module1. Then $2 \times2$ convolution kernels with $stride=1$ are employed to obtain the $256\times6\times6$ feature maps which are then adjusted to the PrimaryCaps layer with 32 convolution 8D capsule channels. Three routing iterations are executed between PrimaryCaps and FaceCaps. Each expression category have a 16D capsule, where each capsule receives inputs from all capsules in the lower layer. We decode it by three fully connected layers, and finally classify the face to one of the emotional categories by squashing function. In the training, Adam algorithm is applied to optimize the network, where learning rate is set to 0.0001. The value of $m^+$ and $m^-$ are 0.9 and 0.1, respectively. The batch size and the maximum iteration are set to 16 and 300, respectively. Our entire E2-Capsnet training is end-to-end. The training of models were completed on a NVIDIA GTX 1080 GPU with 11GB memory. It took about 24 hours to finish optimizing the model.

\section{Experimental Results}
In order to verify the effectiveness of the proposed method, we conduct experiments and evaluations on RAF-DB and EmotioNet. The ablation experiments and analysis of our E2-Capsnet on RAF-DB are provided. Our method is compared with the state-of-the-art FER methods on the two datasets.

In ablation experiments, the VGG16 is used as the baseline and compared with the capsule network \cite{4}(Capsnet), VGG16 with attention mechanism (AVGGnet, which is shown with the purple dashed box in Fig.~\ref{fig:rpn}), the capsule network with rich convolution layers (RCCnet, which is shown with the green dashed box in Fig.~\ref{fig:rpn}) and E2-Capsnet. These networks were trained from scratch in our experiment. Table 1 reports the results of these methods on RAF-DB. 
\begin{table}
	
		\processtable{Accuracy comparison of ablation experiment on RAF-DB.}
		
		{\begin{tabular}{c|c|c|c|c}
			\hline
			Method & Dynamic routing & Rich convolution & Attention map & ACC \\
			\hline
			VGG16  &  & \checkmark &  & 78.14\% \\
			Capsnet  & \checkmark &  &  & 76.12\% \\
			AVGGnet  &  & \checkmark & \checkmark & 79.29\% \\
			RCCnet  & \checkmark & \checkmark &  & 84.78\% \\
			E2-Capsnet  & \checkmark & \checkmark & \checkmark & 85.24\% \\
			\hline
		\end{tabular}}{}

    \label{tab:cgg}
\end{table}

As can be seen from Table 1, E2-Capsnet achieves better performance than others on RAF-DB. Compared with AVGGnet, E2-Capsnet outperforms by a big margin (5.95\%). It is because the model with the dynamic routing can capture relationship between local feature and obtain deep feature representation. This can verify the benefit of dynamic routing. Compared with RCCnet, E2-Capsnet performs better. It is because the model with AU attention mechanism can better capture subtle muscle motions than the model without attention. Attention is introduced to enable the model to acquire active areas that are needed to be focused on. Therefore our E2-Capsnet with the two enhancement modules are beneficial to this task.

\begin{figure}[t]
\centering{\includegraphics[width=80mm]{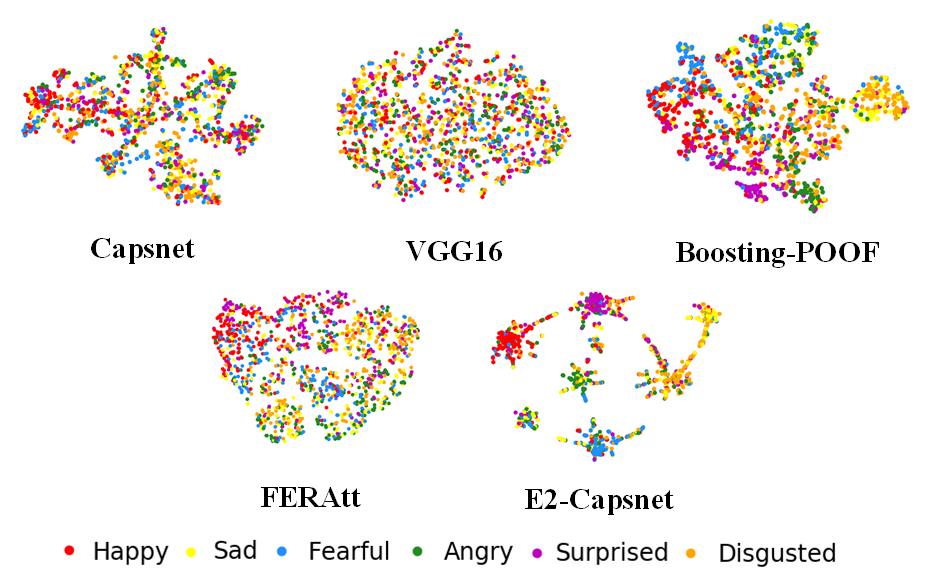}}

\caption{Visualizations of Capsnet \cite{4}, VGG16 \cite{5}, Boosting-POOF \cite{6}, FERAtt \cite{7} and E2-Capsnet on EmotioNet by T-SNE. These points are colored according to their ground-truth class.}  
\label{fig:qwe}
\end{figure}

Table 2 shows the results of our E2-Capsnet compared with the state-of-the-art methods on RAF-DB and EmotioNet. Although most of these methods enhance their learning ability by capturing both the global and local features, our proposed method is superior to all other methods in accuracy as dynamic routing between capsules is employed to obtain the relationship between local features. Capsule networks also encode spatial information while calculating the probability of existence of an object. It is applicable to FER because face has the specific structure. Furthermore, our network can concentrate on active areas of the face enhanced by the attention map. Overall, the proceeds of E2-Capsnet are due to the two enhancement modules, the convolutional neural network with AU-aware attention and the capsule neural network with multiple convolutional layers. In Fig.~\ref{fig:qwe}, the classification results of Capsnet \cite{4}, VGG16 \cite{5}, Boosting-POOF \cite{6}, FERAtt \cite{7} and E2-Capsnet on EmotioNet are also visualized. As shown in Fig.3, our E2-Capsnet can achieve more discriminative representations than the others.
\begin{table}
	\begin{center}
		\processtable{Performance of various FER methods on RAF-DB and EmotioNet.}
		
		{\begin{tabular}{c|c|c|c}
				\hline
				\multirow{2}*{Method} & \multirow{2}*{Source}& \multicolumn{2}{c}{Accuracy}  \\ \cline{3-4}
				
			    & & RAF-DB & EmotioNet \\
				\hline
				
				NCMML \cite{8} & SIP(2016) & 57.70\% & - \\
				Capsnet \cite{4} & arXiv(2017) & 76.12 \% & 32.64\% \\
				Boosting-POOF \cite{6} & FG(2017) & 73.19\% & 46.27\% \\
				MRE-CNN \cite{1} & ICANN(2017) & 76.73\% & - \\
				VGG16 \cite{5} & CS(2014) & 80.96\% & 45.59\% \\
				RC-DLP \cite{9} & CVPR(2017) & 84.70\% & - \\
				Emotion classifier \cite{10} & ICMI(2018) & 80.00\% & - \\
				GAN-Inpainting  \cite{11} & CVPR(2018) & 81.87\% & - \\
			
				DLP-CNN \cite{2} & IEEE TIP(2019) & 84.13\% & - \\
				FERAtt \cite{7} & arXiv(2019) & - & 48.63\% \\
				\textbf {E2-Capsnet} & - & \textbf {85.24}\textbf{\%} & \textbf {55.91}\textbf{\%} \\
				\hline
		\end{tabular}}{}
	\end{center}
	\label{tab:bgg}
\end{table}

\section{Conclusion}
In this paper, we propose an end-to-end E2-Capsnet with AU-aware attention for facial expression recognition. The AU-aware attention can enhance the learning ability of the network in active areas. The capsule neural network with multiple convolutional layers can enhance the ability of the feature representation, which is beneficial for capturing important information in the specific structure of face. The experimental results on two benchmark datasets demonstrate that the proposed E2-Capsnet is superior to the state-of-the-art methods.
\vskip3pt
\ack{This work was supported partly by the Fundamental Research
	Funds for the Central Universities (2019YJS029), the National
	Natural Science Foundation of China (61772067).}

\vskip5pt

\noindent Shan Cao, Yuqian Yao and Gaoyun An  (\textit{Institute of Information Science, Beijing Jiaotong University, Beijing, People's Republic of China})

\vskip3pt

\noindent E-mail: gyan@bjtu.edu.cn

\end{document}